\documentclass[conference]{IEEEtran}
\IEEEoverridecommandlockouts
\usepackage{cite}
\usepackage{amsmath,amssymb,amsfonts}
\usepackage{algorithmic}
\usepackage{graphicx}
\usepackage{textcomp}
\usepackage{xcolor}
\usepackage{float}
\usepackage{booktabs}
\usepackage[lined, boxed, commentsnumbered, ruled]{algorithm2e}
\def\BibTeX{{\rm B\kern-.05em{\sc i\kern-.025em b}\kern-.08em
    T\kern-.1667em\lower.7ex\hbox{E}\kern-.125emX}}
\begin{document}

\title{Towards Accurate Single Panoramic 3D Detection: A Semantic Gaussian Centric Approach}

\author{\IEEEauthorblockN{Kanglin Ning\textsuperscript{1,2}, Yiran Zhao\textsuperscript{1,2}, Wenrui Li\textsuperscript{1}, Shaoru Sun\textsuperscript{1}, Xingtao Wang\textsuperscript{1,2,3}\textsuperscript{*}, Xiaopeng Fan\textsuperscript{1,2,3}}
\IEEEauthorblockA{\textsuperscript{1}Harbin Institute of Technology, \textsuperscript{2}The Suzhou Research Institute of HIT, \textsuperscript{3}The PengChengLab}
\thanks{This work was supported in part by the National Key R\&D Program of China (2025ZD1601300), the National Natural Science Foundation of China (NSFC) under grant 62402138 and 624B2049, the Fundamental and Interdisciplinary Disciplines Breakthrough Plan of the Ministry of Education of China (JYB2025XDXM901), and the Suzhou Key Core Technology Project under grant number SYG2025118.}
\thanks{\textsuperscript{*} Xingtao Wang is the corresponding author.}
}

\maketitle
\begin{abstract}
Three-dimensional object detection in panoramic imagery is crucial for comprehensive scene understanding, yet accurately mapping 2D features to 3D remains a significant challenge. Prevailing methods often project 2D features onto discrete 3D grids, which break geometric continuity and limit representation efficiency. To overcome this limitation, this paper proposes PanoGSDet, a monocular panoramic 3D detection framework built upon continuous semantic 3D Gaussian representations. The proposed framework comprises a panoramic depth estimation component and a semantic Gaussian component. The panoramic depth estimation component extracts the equirectangular semantic and depth features from the monocular panorama input.  The semantic Gaussian component includes a semantic Gaussian lifting module that projects spherical features into 3D semantic Gaussians, a semantic Gaussian optimization module that refines these semantic Gaussians, and a Gaussian guided prediction head that generates 3D bounding boxes from optimized Gaussian representations. Extensive experiments on the Structured3D dataset demonstrate that our method significantly outperforms existing methods.
\end{abstract}


\section{Introduction}
\label{sec:intro}

Panoramic images, with their wide field of view $180^{\circ} \times 360^{\circ}$, enable holistic 3D scene understanding from a single viewpoint. Achieving this understanding fundamentally depends on accurately locating the 3D poses of objects within the scene\cite{lin2025one, wang2025panoextend}.

Early efforts\cite{wang20193d, de2018eliminating, plaut20213d} in single-view panoramic 3D detection primarily adapted mature 2D object detectors by appending an additional pose regression head to predict 3D bounding box attributes directly from 2D proposals.  However, this paradigm was fundamentally constrained by a lack of depth information in the perception process, which limited its accuracy. To address this, subsequent research \cite{zhang2021deeppanocontext, dong2024panocontext, nie2020total3dunderstanding} shifted towards a multi-stage pipeline that first employs an independent monocular depth estimation model to generate a panoramic depth map as an explicit 3D spatial prior. This depth map is subsequently converted into a point cloud to be processed by a point-based 3D detector. However, the point cloud density derived from estimated depth maps is much higher than that measured by 3D scanning equipment such as LiDAR. Therefore, while this addresses the initial spatial-blindness, it introduces a secondary challenge: the standard point cloud sampling and quantization techniques (e.g., FPS, voxelization) required by these detectors\cite{rukhovich2023tr3d, yang2022dbq} often fragment the smooth, continuous surfaces of objects in 3D space, which leads to degraded representation quality and quantization errors.

\begin{figure}[!t]
\centering
\includegraphics[width=3.1in]{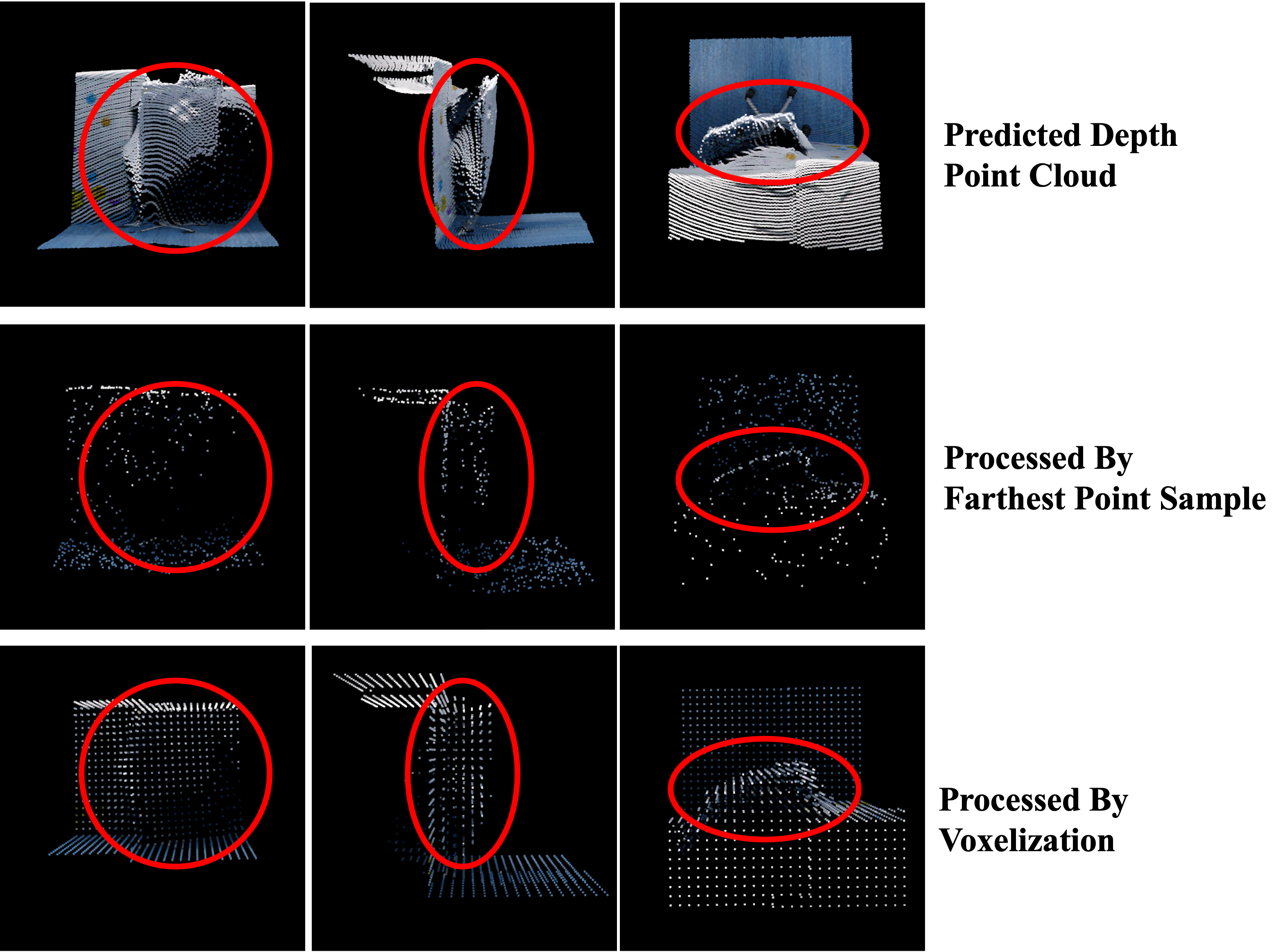}
\caption{A 3D visual comparison between the chair object's original point cloud, farthest point sample, and voxelization of the predicted depth maps by pre-trained Panoformer\cite{shen2022panoformer}. For each processed point cloud, we present images from the front, top, and side views. In the figure, the depth point cloud of the chair object after FPS or voxelization has lost a great deal of its original surface geometry information.}
\end{figure}

Fig. 1 illustrates a representative case where depth estimation for an object (e.g., a chair) is globally accurate but locally inconsistent, exhibiting significant errors in certain regions. Human perception naturally compensates for such incoherence by leveraging strong priors about object surface geometry. In contrast, standard point-based 3D detectors process the derived point cloud through operations like farthest point sampling (FPS) or voxelization. As visualized, these sampling strategies actively disrupt the intrinsic surface continuity of objects, thereby creating a fundamental bottleneck for accurate detection.

\begin{figure*}[!t]
\centering
\includegraphics[width=6.0in]{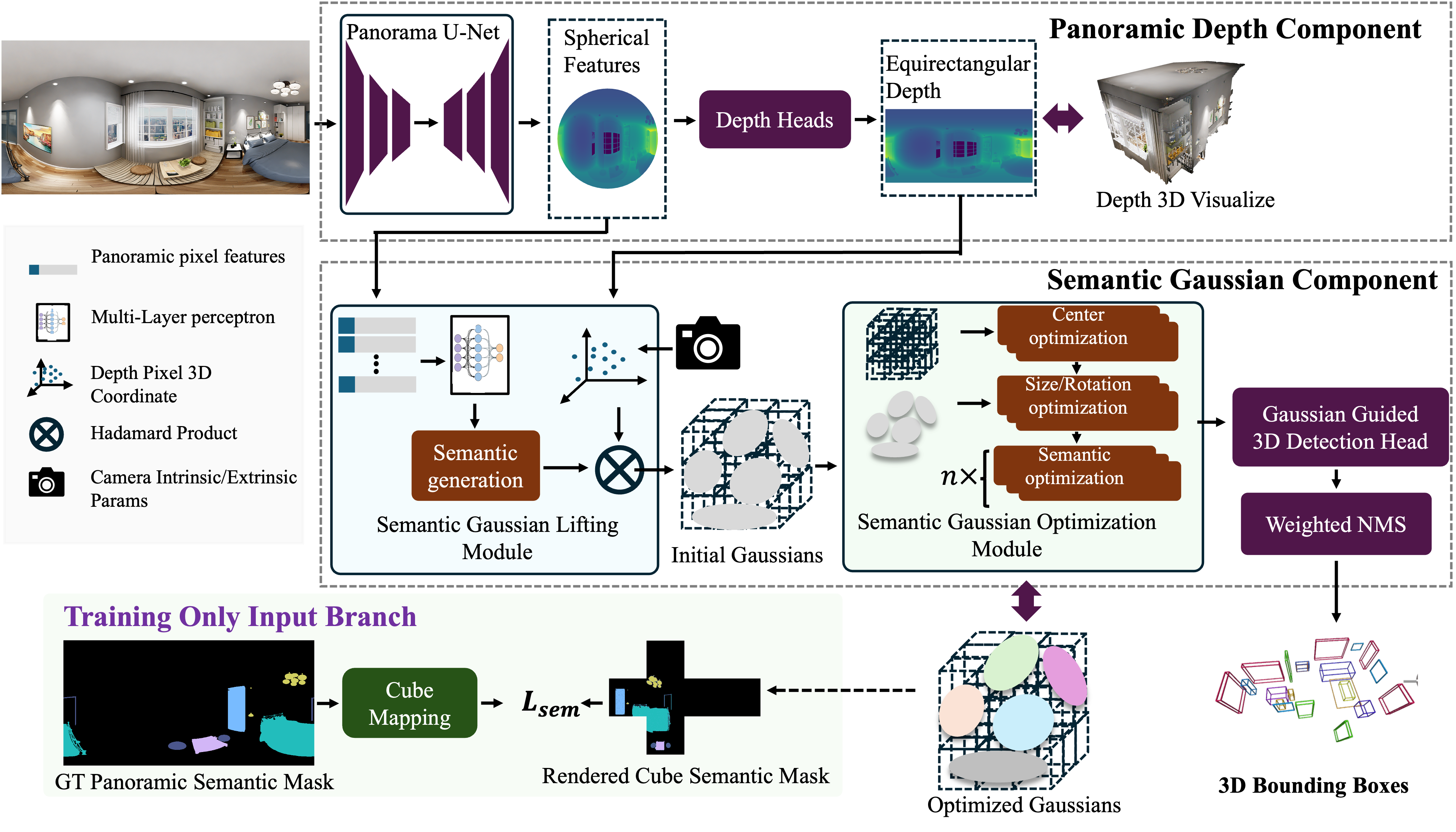}
\caption{The pipeline of our proposed 3D object detector PanoGSDet. The detector comprises a depth estimation branch and a detection branch. The detection branch uses multi-scale features extracted by the depth estimation branch and the depth map, combined with camera intrinsic and extrinsic parameters, to map them into 3D space as semantic gaussian represents.}
\end{figure*}

To address the geometric continuity limitations inherent in existing representation schemes, we propose PanoGSDet, a monocular panoramic 3D object detection framework based on a continuous semantic 3D Gaussian representation. Our framework comprises two core components: a panoramic depth estimation component and a semantic 3D Gaussian component. The depth estimation component extracts spherical semantic features and depth information from the input panorama. The semantic Gaussian component includes a semantic Gaussian lifting module, a semantic Gaussian optimization module, and a Gaussian guided 3D detection head. The semantic Gaussian lifting module uses the estimated depth map combined with the spherical semantic feature vectors of corresponding pixels to predict the initial position, size, rotation angle, opacity, and semantic category of each semantic 3D Gaussian ball. Considering that inherent depth estimation errors, we refine this representation through a semantic Gaussian optimization module that adjusts Gaussian parameters in position, scale, rotation, and semantic information. To supervise this optimization, the proposed optimization module renders semantic 3D Gaussians into cube-map semantic segmentation maps and supervises them with a semantic segmentation loss. Finally, a Gaussian-guided 3D detection head generates precise 3D bounding boxes from the optimized foreground semantic 3D Gaussian balls. To verify the performance of our method, we conducted extensive experiments on the open-source dataset Structured3D, and the results show that our method has a significant performance advantage compared to existing strategies.

Our contribution can be summarized as follows:
\vspace{-\topsep}
\begin{itemize}
\item We proposed a novel monocular panoramic 3D detection framework named PanoGSDet. The proposed framework mitigates the geometric discontinuity by introducing a continuous semantic 3D Gaussian representation.
\item We proposed a semantic 3D Gaussian lifting module. The proposed lifting module uses the depth and semantic feature of corresponding pixels to predict the initial semantic 3D Gaussian representation.
\item We proposed a semantic 3D Gaussian optimization module. The proposed optimization module renders semantic 3D Gaussians into cube-map semantic segmentation maps and supervises them with a segmentation loss against the ground-truth cube maps.
\end{itemize}

\section{Method}

The framework of our proposed PanoGSDet is shown in Fig. 2. This framework includes a panoramic depth estimation component and a semantic 3D Gaussian component. For the panoramic depth estimation component, the proposed framework adopt the current state-of-the-art depth estimate model, Panoformer \cite{shen2022panoformer}.  To ensure the reliability of depth estimation predictions during training, we initialize the depth component with pre-trained weights on the Structured3D dataset and freeze these weights during training. With relatively accurate depth and semantic information, our semantic Gaussian lifting module predicts the corresponding initial Gaussian balls from the feature vector of each feature pixel. The semantic Gaussian optimize module optimizes the properties of each Gaussian ball to mitigate the influence of inherent errors in the depth estimation module. Finally, the Gaussian-guided 3D detection head predicts the corresponding 3D bounding boxes from each foreground Gaussian ball.

\subsection{Semantic 3D Gaussian Lifting Module}
Our semantic 3D Gaussian representation extends the conventional geometric parameterization (center coordinates, covariance matrix, opacity) with two key semantic attributes: a category label and a  semantic feature vector. To fully exploit this enriched representation, the semantic Gaussian lifting module is designed to directly predict the initial parameters of each Gaussian from the provided 2D spherical semantic and depth information.

To initialize the approximate 3D location for each semantic Gaussian prior to constructing the full representation, we directly leverage the spherical depth map. Specifically, given a depth map $D \in \mathbf{R}^{B \times H \times W}$, where $B, H, W$ denote the batch size, height and width of input equirectangular panorama, the 3D coordinates of a point corresponding to pixel $p = \{ d(u, v) | u = 1, ..., H; v = 1, ..., W; \}$ are obtained by transforming its 2D position and depth value into a camera-centric 3D coordinate system using the following transformation:

\begin{equation}
\begin{aligned}
& \theta = (\frac{u}{H} - 0.5) \times \pi , \eta = (1 -\frac{v}{W}) \times 2\pi \\
& x = d(u,v)\sin (\eta) \\
& y = d(u,v)\cos(\theta)\cos(\eta) \\
& z = d(u,v)\cos(\eta)\sin(\theta),
\end{aligned}
\end{equation}
where $d(u,v)$ represents the depth value of the pixel in row $u$ and column $v$ in the depth map $D$. For given depth map $D$, the set of 3D points derived from all pixels is represented as $C = \{  (x_j, y_j, z_j) | j= 1,..., H \times W\}$. 

Based on the obtained 3D point set, we instantiate a collection of semantic 3D Gaussians. Each Gaussian is centered at a 3D point from this set, and its associated semantic feature vector is sourced from the corresponding location in the panoramic feature map. The remaining attributes—scale, rotation, opacity, and semantic category—are predicted from this feature vector via a lightweight MLP network. The opacity and category represent semantic-level properties; their MLP outputs are constrained to the range $[0, 1]$ using a sigmoid activation. 

The scale and rotation matrices define the ellipsoidal geometry of each Gaussian. Since the scale matrix is a $3 \times 3$ diagonal matrix, it can be physically interpreted as a triple of radii $r_i = (r_x, r_y, r_z)$ along the principal axes. Similarly, the rotation matrix is decomposed into Euler angles $\theta_i = (\theta_x, \theta_y, \theta_z)$. Thus, initializing these matrices simplifies to predicting a scale triplet and a rotation triplet. To maintain physical plausibility, each rotation angle is constrained to $[-\pi, \pi]$. The prediction for the scale triplet is formulated as:

\begin{equation}
\begin{aligned}
& r_i = N(\Gamma(f_i)) \times R_{max} \\
& \theta_i =(N(\Gamma(f_i)) - 0.5) \times 2 \times \pi,
\end{aligned}
\end{equation}
where $\Gamma$ represents the mapping function composed of MLP networks, $R_{max}$ represents the maximum radius set by us, $f_i$ represents the corresponding semantic feature vector, and $N$ denotes the sigmoid function.

\subsection{Semantic Gaussian Optimization Module}

Considering that the initial 3D semantic Gaussians are derived from potentially noisy depth estimates and predicted attributes (scale, rotation, opacity, and category), their parameters may be inaccurate. To address this, our semantic Gaussian optimization module refines each Gaussian across three key dimensions: position (center), geometric structure (covariance), and semantic properties. This optimization is performed progressively through a series of $n$ stacked sub-modules. In each sub-module, updates are applied sequentially—first to semantic information, then to the center position, and finally to the covariance matrix. This sequence ensures stable and effective refinement of the Gaussian representation.

\textbf{1) Semantic optimization}

When optimizing the semantic category and feature vector of a 3D semantic Gaussian, the information within an individual Gaussian is often insufficient. To refine the semantic features within a local region and expand the receptive field from a single Gaussian to its neighborhood, we draw inspiration from established point cloud feature processing techniques. For computational efficiency and implementation practicality, we adapt and employ sparse 3D convolution. This allows joint optimization of the semantic category and feature vector for each 3D semantic Gaussian.

Given all Gaussians in the scene, we discretize them into a 3D voxel feature volume based on their center coordinates and corresponding semantic feature vectors. To avoid merging neighboring points from the same Gaussian, we use a sufficiently small voxel grid size during voxelization. The resulting discrete voxel feature volume is processed by two consecutive blocks. Each block consists of a Sparse SubMConv3d \cite{yan2018second} layer with a $5 \times 5 \times 5$ kernel, followed by LayerNorm and GELU. This setup refines the feature vector for each Gaussian. For semantic category optimization, a separate Sparse SubMConv3d layer with a $1 \times 1 \times 1$ kernel predicts a new category distribution. To ensure gradual refinement, the updated category scores are averaged with the original category distribution $C_o$ to produce the final category distribution $C_p$. This process can be summarized as follows:

\begin{equation}
C_n = (Softmax(\Gamma_{3d}(f_i)) + C_o) / 2,
\end{equation}
where $\Gamma_{3d}$ is a mapping function consisted of Sparse SubMConv3d layers, $f_i$ denotes the corresponding feature vector.

\begin{table*}[!t]
\scriptsize
\caption{Comparison with current state of the art with metric mAP@25.}
\centering
\begin{tabular}{lcccccccccccc}
\toprule[1.2pt]
Methods            & bed    & chair  & sofa   & table  & desk   & dresser & cabinet & fridge & sink   & lamp   & bathtub & mAP@25 \\ \midrule[1.2pt]
MSCNN\cite{wang20193d} & 0.6123 & 0.0423 & 0.0531 & 0.1124 & 0.1613  & 0.2011& 0.1833 & 0.2021 & 0.0799 & 0.0421 & 0.4531 & 0.1948 \\
EBS\cite{de2018eliminating} & 0.6233 & 0.0512 & 0.0566 & 0.1211 & 0.1912 & 0.2113 & 0.1912 & 0.2101 & 0.0891 & 0.0512 & 0.4673 & 0.2057 \\
Total-Pano\cite{nie2020total3dunderstanding} & 0.7762 & 0.0607 & 0.0852 & 0.1672 & 0.1943 & 0.2870  & 0.2327  & 0.2614 & 0.1414 & 0.1145 & 0.5194  & 0.2367 \\
Im3D-Pano\cite{zhang2021holistic} & 0.6242 & 0.0999 & 0.0723 & 0.2194 & 0.2540 & 0.4379  & 0.2411  & 0.3909 & 0.0961 & 0.1225 & 0.5116  & 0.2790 \\
DeepPanoContext\cite{zhang2021deeppanocontext} & 0.7001 & 0.0644 & 0.0644 & 0.1675 & 0.2189 & 0.4233  & 0.2501  & 0.4209 & 0.1500 & 0.1537 & 0.6724  & 0.2989 \\
PanoContext-Former\cite{dong2024panocontext} & 0.8588 & 0.1144 & 0.1259 & 0.2599 & 0.5252 & 0.1852  & 0.1756  & 0.6221 & 0.6221 & 0.6221 & 0.6221  & 0.3680 \\
TR3D-PanoFormer\cite{rukhovich2023tr3d} & 0.8600 & 0.1146 & 0.1261 & 0.2601 & 0.3397 & 0.4956  & 0.3478 & 0.5254 & 0.1854 & 0.1758 & 0.6222  & 0.3682 \\
Ours                           & \textbf{0.9023} & \textbf{0.2314} & \textbf{0.2142} & \textbf{0.3033} & \textbf{0.4215} & \textbf{0.5234} &  \textbf{0.4071} & \textbf{0.5854} & \textbf{0.2562} & \textbf{0.2344} & \textbf{0.6745} & \textbf{0.4321} \\ \bottomrule[1.2pt]
\end{tabular}
\end{table*}

\begin{table*}[!t]
\scriptsize
\caption{Comparison with current state of the art with metric mAP@50.}
\centering
\begin{tabular}{lcccccccccccc}
\toprule[1.2pt]
Methods            & bed    & chair  & sofa   & table  & desk   & dresser & cabinet & fridge & sink   & lamp   & bathtub & mAP@50 \\ \midrule[1.2pt]
MSCNN\cite{wang20193d} & 0.4712 & 0.0092 & 0.0089 & 0.0213 & 0.0091 & 0.0731 & 0.0542 & 0.1121 & 0.0123 & 0.0111 & 0.1514 & 0.0849\\
EBS\cite{de2018eliminating} & 0.4812 & 0.0101 & 0.0103 & 0.0311 & 0.0121 & 0.0832 & 0.0711 & 0.1213 & 0.0231 & 0.0224 & 0.1818 & 0.0952 \\
Total-Pano\cite{nie2020total3dunderstanding} & 0.5416 & 0.0135 & 0.0238 & 0.0484 & 0.0152 & 0.0903  & 0.0798  & 0.1491 & 0.0316 & 0.0242 & 0.2913  & 0.1090 \\
Im3D-Pano\cite{zhang2021holistic}  & 0.4840 & 0.0278 & 0.0262 & 0.0866 & 0.0835 & 0.1692  & 0.1309  & 0.3314 & 0.0446 & 0.0364 & 0.1815  & 0.1456 \\
DeepPanoContext\cite{zhang2021deeppanocontext} & 0.4921 & 0.0191 & 0.0230 & 0.0566 & 0.0512 & 0.1424  & 0.0994  & 0.2570 & 0.0500 & 0.0381 & 0.4431  & 0.1520 \\
PanoContext-Former\cite{dong2024panocontext} & 0.7157 & 0.0377 & 0.0569 & 0.0962 & 0.1382 & 0.1688  & 0.2081  & 0.4703 & 0.0676 & 0.0590 & 0.2845  & 0.2093 \\
TR3D-PanoFormer\cite{rukhovich2023tr3d} & 0.6115 & 0.0335 & 0.0428 & 0.0766 & 0.0826 & 0.1240  & 0.1560  & 0.4063 & 0.0742 & 0.0403 & 0.4906 & 0.1944 \\
Ours               & 0.7040 & \textbf{0.0539} & \textbf{0.0753} & \textbf{0.1276} & \textbf{0.1429} & \textbf{0.2270} & 0.2012 & 0.4238 & \textbf{0.1071} & \textbf{0.0624} & \textbf{0.5829} & \textbf{0.2461} \\ \bottomrule[1.2pt]
\end{tabular}
\end{table*}

\textbf{2) Center Optimize}

When optimizing the center point of a semantic 3D Gaussian, our objective is to better align its surface with the underlying object geometry. Each center refinement block takes the semantic feature vector of the corresponding Gaussian as input and predicts a 3D offset $idx = (\triangle x, \triangle y, \triangle z)$ toward the object’s true surface. This offset is then added to the current center position via residual addition, shifting the Gaussian a small step toward the target location. The update for each center point $c_i = (x_i, y_i, z_i)$ within one optimization block is expressed as follows:
\begin{equation}
\begin{aligned}
& idx = (N(MLP(f_i)) - 0.5) \times S      \\
& \hat{c_i} = c_i + idx,
\end{aligned}
\end{equation}
where $f_i$﻿  denotes the semantic feature vector of the Gaussian, $MLP$ is a lightweight mapping network, and $N$ denotes a normalization function that constrains the MLP output to the range $(0, 1)$. The scale hyper-parameter $S$﻿ controls the maximum allowable offset per optimization step, and $\hat{c_i}$ represents the updated center coordinates of the Gaussian after refinement.

\textbf{3) Covariance Matrix Optimization}

To preserve the positive semi-definite property of the covariance matrix during optimization, we follow the practice in the original 3D Gaussian Splatting (3DGS). We decompose the matrix into a scale matrix and a rotation matrix for separate refinement. Thus, optimizing the covariance matrix for each Gaussian is equivalent to optimizing its scale triple $\mathbf{r}_i = (r_x, r_y, r_z)$ and rotation triple $\boldsymbol{\theta}_i = (\theta_x, \theta_y, \theta_z)$. These parameters correspond to physically meaningful quantities (radii and Euler angles), so their optimization is constrained within plausible ranges. Using the residual paradigm from center refinement, we also optimize these six parameters incrementally. The semantic feature vector of each Gaussian is used to predict an offset vector $(sc_i, ro_i)$ via a lightweight network. This offset vector is then added to the current values to get the refined parameters. This process is formalized as follows:

\begin{equation}
\begin{aligned}
& sc_i = (N(MLP(f_i)) - 0.5) \times \beta \\
& ro_i = (N(MLP(f_i)) - 0.5) \times \eta \\
& \mathbf{\hat{r_i}} = (\mathbf{r_i} +sc_i) / 2 \\
& \mathbf{\hat{\theta_i}} = (\mathbf{\theta_i} + ro_i) / 2,
\end{aligned}
\end{equation}

where $\beta, \eta$ are two hyper-parameter to control the optimize range of $r_i$ and $\theta_i$, $\hat{r_i}$ and $\hat{\theta_i}$ are the optimized scale and rotation triplet tuple.

\subsection{Gaussian Guided 3D Detection Head}

After obtaining the refined semantic 3D Gaussian representation of the scene, the final stage involves predicting the corresponding 3D bounding boxes. The prediction is directly generated from the feature vector of each Gaussian. A distinct advantage of our representation is the inherent availability of semantic labels: unlike detectors built upon other 3D representations, we can directly adopt the semantic category of each Gaussian as the initial class label for its associated box. Consequently, for each Gaussian, the detection head only needs to regress the bounding box center, size, rotation, and a confidence score. To improve efficiency, we first filter the Gaussians by their semantic categories, retaining only those belonging to foreground objects. For the regression and confidence prediction head, we adopt the design paradigm of TR3D. Specifically, for each remaining Gaussian, we predict an offset from its center to the final bounding box center, while the size and rotation of the box are regressed following the standard formulation of TR3D.

\begin{figure*}[!t]
\centering
\includegraphics[width=6.5in]{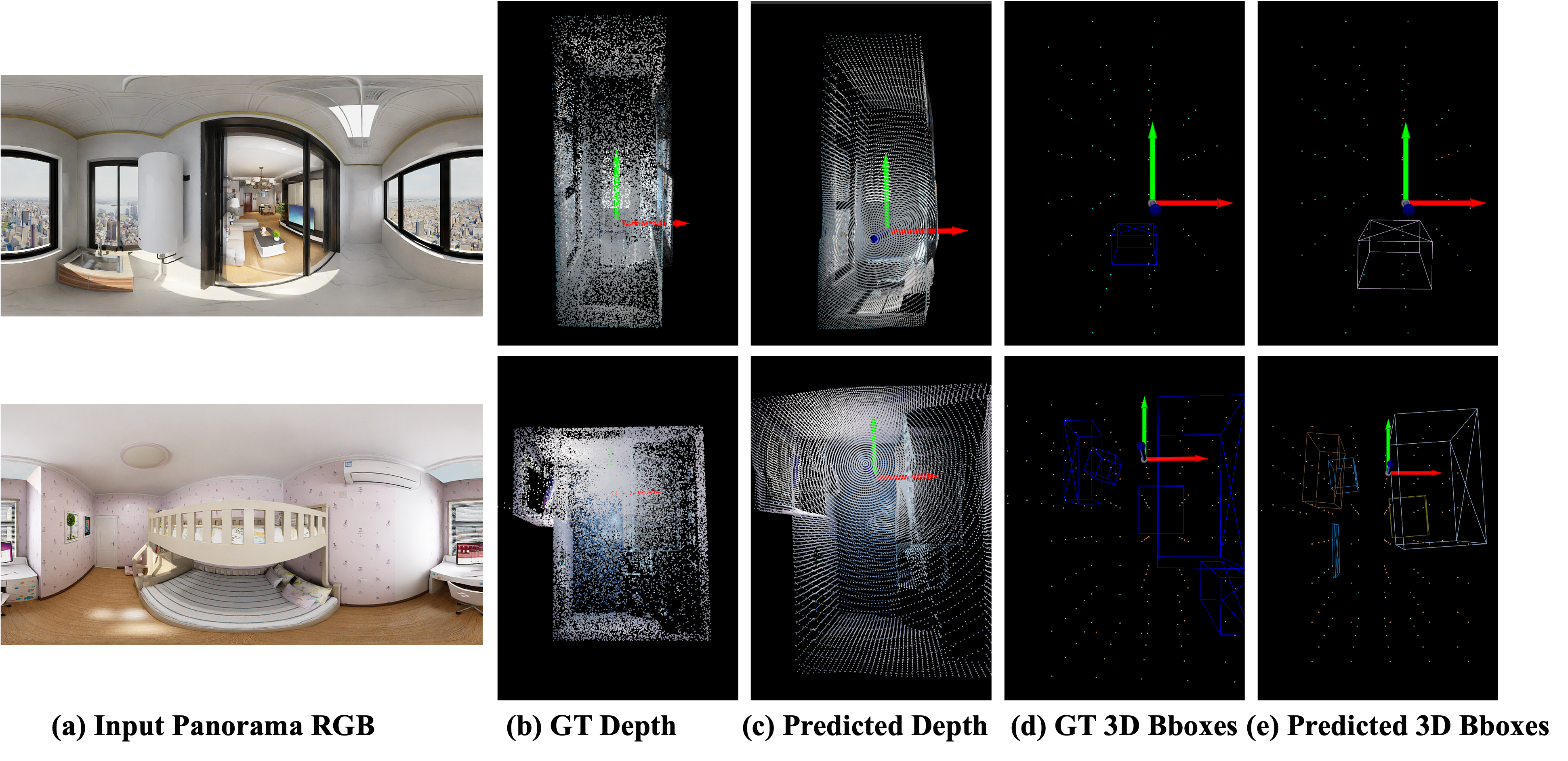}
\caption{The qualitative analysis of our proposed PanoGSDet. The first column shows the input panoramic RGB images. The second column displays the 3D visualization of the ground-truth depth maps. The third column presents the 3D visualization of the depth estimation branch prediction results. The fourth column shows the visualization of ground-truth 3D bounding boxes. The fifth column presents the PanoGSDet's prediction results.}
\vspace{-0.5cm}
\end{figure*}

\subsection{Objective Function}
To supervise the learnable optimization blocks in our semantic 3D Gaussian module, we use the ground-truth panoramic semantic mask as explicit guidance. During training, we render the optimized Gaussians onto the six faces of a cube map (up, down, left, right, front, back). At the same time, we project the ground-truth panoramic semantic mask onto an identical cube map to obtain per-face targets. We compute the cross-entropy loss between each rendered view and its corresponding target for supervision. Formally, let the rendered cube map set be $\mathcal{M} = \{M_{\text{up}}, M_{\text{down}}, M_{\text{front}}, M_{\text{back}}, M_{\text{left}}, M_{\text{right}}\}$ and the ground-truth cube map be $\mathcal{G} = \{G_{\text{up}}, G_{\text{down}}, G_{\text{front}}, G_{\text{back}}, G_{\text{left}}, G_{\text{right}}\}$. The objective function for semantic optimization is:
\begin{equation}
L_{sem} =\frac{1}{6} \sum_{m\in M, g \in G} BCE(m, g),
\end{equation}
In addition to the semantic optimization objective, our 3D detection is jointly supervised by a regression loss, $L_{reg}$, and a confidence loss, $L_{conf}$. For both terms, we follow the formulation used in TR3D. Thus, the complete objective function can be expressed as:
\begin{equation}
L = L_{sem} + L_{reg} + L_{conf},
\end{equation}

\section{Experiments}

\subsection{Experimental Settings}
\textbf{Dataset:} We utilized the Structured3D dataset \cite{zheng2020structured3d} to conduct our experiments. This dataset contains 196K rendered panoramic images and corresponding depth maps, comprising 3500 scenes and 12835 rooms. The processed dataset comprises 11 categories of 3D bounding boxes, each labeled with a corresponding rotation angle. Each room has three modes: Normal, Full, and Empty. Considering the requirements of our detection task, we utilized the depth map annotations from the full mode (which contains the most 3D bounding boxes) to generate point clouds. Following the official approach, we used the first 3000 scenes from the entire dataset as the training set, scenes 3000-3250 as the validation set, and the last 250 scenes as the test set.

\textbf{Metric:} Following the approach adopted by\cite{rukhovich2023tr3d} et al., we used  AP@25 and AP@50 as the main evaluation metrics. AP@25 indicates that an IoU threshold of 0.25 is used to determine whether a bounding box is a positive sample, and AP@50 uses an IoU threshold of 0.5 to determine whether a bounding box is positive or negative.

\textbf{Training Settings:} Throughout the training process, we used AdamW as the optimizer, setting its learning rate to 0.001, weight decay to 0.001, and gradient clipping threshold to 35. We trained the model for 12 epochs on the Structured3D dataset, employing a multi-step learning rate schedule strategy, setting the learning rate decay steps to the 8th and 11th epochs, respectively.


\begin{table}
\caption{Training and Inference Consumption Comparison. The values in parentheses indicate the video memory used by the depth estimation part.}
\scriptsize
\centering
\begin{tabular}{lccc}
\toprule[1.2pt]
Methods & GPU Memory Training & GPU Memory Testing & FPS \\
\midrule[1.2pt]
DeepPanoContext\cite{zhang2021deeppanocontext} & 22145 (2876) & 8653(2876) & 1 \\
PanoContextFormer\cite{dong2024panocontext} & 21122(2876) & 8342(2876) & 2 \\
Ours &  3893 (2876) & 3197(2876) & 11 \\
\bottomrule[1.2pt]
\end{tabular}
\end{table}

\subsection{Comparison with current state of the art}

\begin{table}
\caption{Effectiveness of each component. }
\scriptsize
\centering
\begin{tabular}{lccc}
\toprule[1.2pt]
Gaussian Lifting & Gaussian Optimize & mAP@25 & mAP@50 \\
\midrule[1.2pt]
 & & 0.3682 & 0.1944 \\
 \checkmark & & 0.3852 & 0.2223 \\
 & \checkmark &  0.3922 & 0.2112  \\
\checkmark & \checkmark & \textbf{0.4321} & \textbf{0.2461} \\
\bottomrule[1.2pt]
\end{tabular}
\end{table}

The experimental results on the Structured3D dataset are shown in Tables  \uppercase\expandafter{\romannumeral1}, \uppercase\expandafter{\romannumeral2} and  \uppercase\expandafter{\romannumeral3}. Tables  \uppercase\expandafter{\romannumeral1} and  \uppercase\expandafter{\romannumeral2} primarily compare the accuracy of our method with that of other current methods. In these experiments, we compared our method with Deeppanocontext\cite{zhang2021deeppanocontext}, Total-pano\cite{nie2020total3dunderstanding}, Im3D-Pano\cite{zhang2021holistic}, and Panocontext-former\cite{dong2024panocontext}. Table  \uppercase\expandafter{\romannumeral3} shows the differences in computational resource usage between our method and image-based methods, including GPU memory usage and frequency per second during training. 

As shown in Table \uppercase\expandafter{\romannumeral1} and \uppercase\expandafter{\romannumeral2}, our method significantly outperforms existing panorama-based methods in terms of mAP@25 and mAP@50. Furthermore, compared with TR3D trained using point clouds generated by the same method that uses Panoformer to predict depth maps, our method also demonstrates superior performance. This demonstrates that, compared to Panocontext-former's approach of purely predicting decoded point clouds from depth maps, our method's introduction of a semantic 3D Gaussian is essential.

As shown in Table  \uppercase\expandafter{\romannumeral3}, our method is highly competitive in terms of computational resource consumption. Similar to deep panocontext and panocontext former, our method freezes the weights of the depth estimation model during training, thus significantly reducing the computational resource consumption of the depth estimation model. During this process, the GPU memory consumption for the depth estimation part per panorama image is reduced to approximately 2.8 GB. Correspondingly, the GPU memory consumption for the 3D semantic Gaussian component, when processing a single panorama image during training, is approximately 1 GB, whereas the consumption for this part of the network during inference is only 321 MB. In terms of frequency per second, our method also has a significant advantage over Deeppanocontext and Panocontext-former.
 
\subsection{Qualitative Analysis}

To further validate the effectiveness of our method, we visualized its 3D detection performance on the Structured3D dataset. The visualization of experiment results, shown in Fig. 3, includes the original input RGB panorama images, the 3D visualization of the ground-truth depth maps, the 3D visualization of the predicted depth maps, the ground-truth 3D bounding boxes, and the predicted 3D bounding boxes. The results show that our method achieves very accurate detection for large objects. In future work, we plan to investigate the use of the clip paradigm to assist the detector in better identifying these target categories and explore open-set detection tasks.

\subsection{Ablation Study}

In this section, we conduct ablation experiments to validate the effectiveness of our method. Our baseline is TR3D trained using depth point clouds predicted by Panoformer, and we progressively add our proposed method for comparative experiments. The results show that each of our proposed methods delivers a significant performance improvement. Interestingly, while our semantic optimization module works effectively, its performance improvement is not as pronounced as our semantic Gaussian lifting strategy. This illustrates that an efficient 3D representation is essential for 3D perception tasks.

\section{Conclusion}
This paper proposes a monocular panoramic 3D detection framework built on 3D Gaussian representations named as PanoGSDet. The proposed framework comprises a panoramic depth estimation component and a semantic Gaussian component. The panoramic depth estimation component extracts the equirectangular semantic and depth features from monocular panorama input. The semantic Gaussian component includes a semantic Gaussian lifting module, a semantic Gaussian optimization module, and a Gaussian-guided 3D detection head.  An extensive experiment on the Structured3D dataset demonstrates that our method achieves superior performance in the 3D object detection task.

\bibliographystyle{IEEEbib}
\bibliography{icme2026references}

\vspace{12pt}

\end{document}